\documentclass[10pt,twocolumn,letterpaper]{article}

\usepackage{iccv}
\usepackage{times}
\usepackage{epsfig}
\usepackage{graphicx}
\usepackage{amsmath}
\usepackage{amssymb}

\usepackage{multirow}
\usepackage{booktabs}
\usepackage{array}
\usepackage{tabularx}
\usepackage{tabulary}
\usepackage{makecell}
\usepackage{color, colortbl}
\usepackage{tabu}
\usepackage{xcolor}
\usepackage{float}
\usepackage{graphicx}
\usepackage{caption}

\usepackage[breaklinks=true,bookmarks=false]{hyperref}

\iccvfinalcopy 

\def\iccvPaperID{****} 

\ificcvfinal\fi

\newcommand{\cocadraw}{CoBIT}
\newcommand{\cocadrawbase}{CoBIT-Base}
\newcommand{\cocadrawlarge}{CoBIT-Large}
\newcommand{\bfcocadraw}{\textbf{CoBIT}}

\newcommand{\red}[1]{\underline{#1}}
\newcommand{\redbf}[1]{\textbf{\underline{#1}}}
\definecolor{lg}{gray}{0.9}
\definecolor{lb}{gray}{0.35}
\definecolor{llb}{gray}{0.6}
\newcommand{\lb}[1]{\textcolor{lb}{ #1}}
\newcommand{\llb}[1]{\textcolor{llb}{ #1}}
\newcommand\blfootnote[1]{%
\begingroup
\renewcommand\thefootnote{}\footnote{#1}%
\addtocounter{footnote}{-1}%
\endgroup
}

\title{\cocadraw: A Contrastive Bi-directional Image-Text Generation Model}
\author{{\fontsize{11}{10} \selectfont Haoxuan You$^{1*}$,  Mandy Guo$^2$, Zhecan Wang$^1$, Kai-Wei Chang$^3$, Jason Baldridge$^2$, Jiahui Yu$^2$}   \\
{\fontsize{11}{10} \selectfont $^1$Columbia University}\\
{\fontsize{11}{10} \selectfont $^2$Google Research}\\
{\fontsize{11}{10} \selectfont $^3$UCLA}\\
{\tt\small haoxuan.you@cs.columbia.edu, jiahuiyu@google.com}
}
\begin{document}
\twocolumn[{%
\renewcommand\twocolumn[1][]{#1}%
\maketitle
\begin{center}
    \centering
    \captionsetup{type=figure}
    \includegraphics[width=1.03\linewidth]{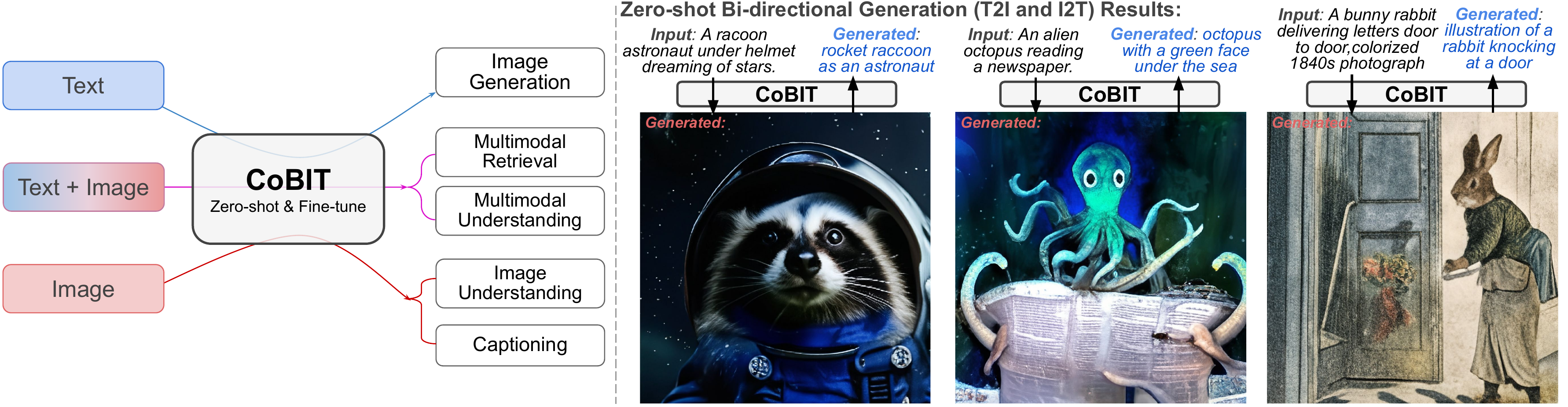}
    \vspace{-6mm}
    \captionof{figure}{We propose \cocadraw{}, a unicoder-decoder architecture, pre-trained jointly on contrastive loss, image-to-text generation loss and text-to-image generation loss. \cocadraw{}  can address a variety of vision and vision-language tasks 
    in the manner of both zero-shot and fine-tuning. The right-hand side displays the zero-shot generated images by \cocadraw{} given novel prompts, and the zero-shot generated captions by \cocadraw{} given the previously generated images as input. }
    \label{fig:intro}
\end{center}%
}]



\maketitle
\ificcvfinal\thispagestyle{empty}\fi

\blfootnote{$^*$ This work was done when Haoxuan was an intern at Google. }

\begin{abstract}
\vspace{-4mm}
  The field of vision and language has witnessed a proliferation of pre-trained foundation models. Most existing methods  are independently pre-trained with contrastive objective like CLIP, image-to-text generative objective like PaLI, or text-to-image generative objective like Parti. However, the three objectives can be pre-trained on the same data, image-text pairs, and intuitively they complement each other as contrasting provides global alignment capacity and generation grants fine-grained understanding. In this work, we present a \textbf{Co}ntrastive \textbf{B}i-directional \textbf{I}mage-\textbf{T}ext generation model (\bfcocadraw), which attempts to unify the three pre-training objectives in one framework.   Specifically, \cocadraw{} employs a novel unicoder-decoder structure, consisting of an image unicoder, a text unicoder and a cross-modal decoder. The image/text unicoders can switch between encoding and decoding in different tasks, enabling flexibility and shared knowledge that benefits both image-to-text and text-to-image generations. \cocadraw{} achieves superior performance in image understanding, image-text understanding (Retrieval, Captioning, VQA, SNLI-VE) and text-based content creation, particularly in zero-shot scenarios. For instance, 82.7\% in zero-shot ImageNet classification, 9.37 FID score in zero-shot text-to-image generation and 44.8 CIDEr in zero-shot captioning.

   
\end{abstract}

\section{Introduction}


For vision-and-language learning, foundation model development has been primarily dedicated to the following directions. (1)  Guiding visual representation pre-training using correlated textual descriptions, using contrastive losses ~\cite{radford2021learning, yao2021filip, mu2022slip, you2022learning}. In this line of work, two unimodal encoders encode image and text, respectively. Such pre-trained \textit{dual-encoder models} can support cross-modal matching downstream tasks, including zero-shot image classification, image-text retrieval, and more.  Also, the visual encoders exhibit strong representational capacity for core image processing tasks, especially classification. (2) Pre-training \textit{image-encoder \& text-decoder models} with Image-to-Text  (I2T) token generation loss \cite{wang2022git, chen2022pali, wang2021simvlm, alayrac2022flamingo}. Pre-trained generative models of this form have demonstrated strong capabilities in vision-and-language tasks such as Visual Question Answering (citation needed--Agrawal et al, IIRC) and image captioning. (3) Pre-training \textit{text-encoder \& image-decoder models} with Text-to-Image (T2I) (visual) token generation loss \cite{ramesh2021zero, yu2022scaling, chang2022maskgit, chang2023muse}. In such work, raw images are tokenized/quantized into a sequence of image tokens by VQ-VAE/GAN models \cite{NIPS2017_7a98af17, yu2021vector}, such that the text-to-image generation problem can be formulated as a standard sequence-to-sequence task. 
The resulting models have shown strong results in downstream text-based image generation tasks.

Most of above-mentioned models are trained independently with only one of these three objectives. Only a few are trained with a combination of two objectives; e.g. CoCa ~\cite{yu2022coca} combines contrastive learning and image-to-text generation. OFA \cite{wang2022ofa} and UnifiedIO \cite{lu2022unified} ensemble image-to-text generation and text-to-image generation. However, it's worth noting that the three objectives share the same pre-training data source: image-text pairs. Intuitively, the knowledge they learn should complement each other: for example, contrastive learning drives high-level multimodal matching, whereas image/text generation requires more fine-grained image-text representations. Last but not least, joint pre-training enables partially sharing of the computational graphs, and thus can be optimized and deployed more efficiently.

\begin{figure*}[t]
  \includegraphics[width=\linewidth]{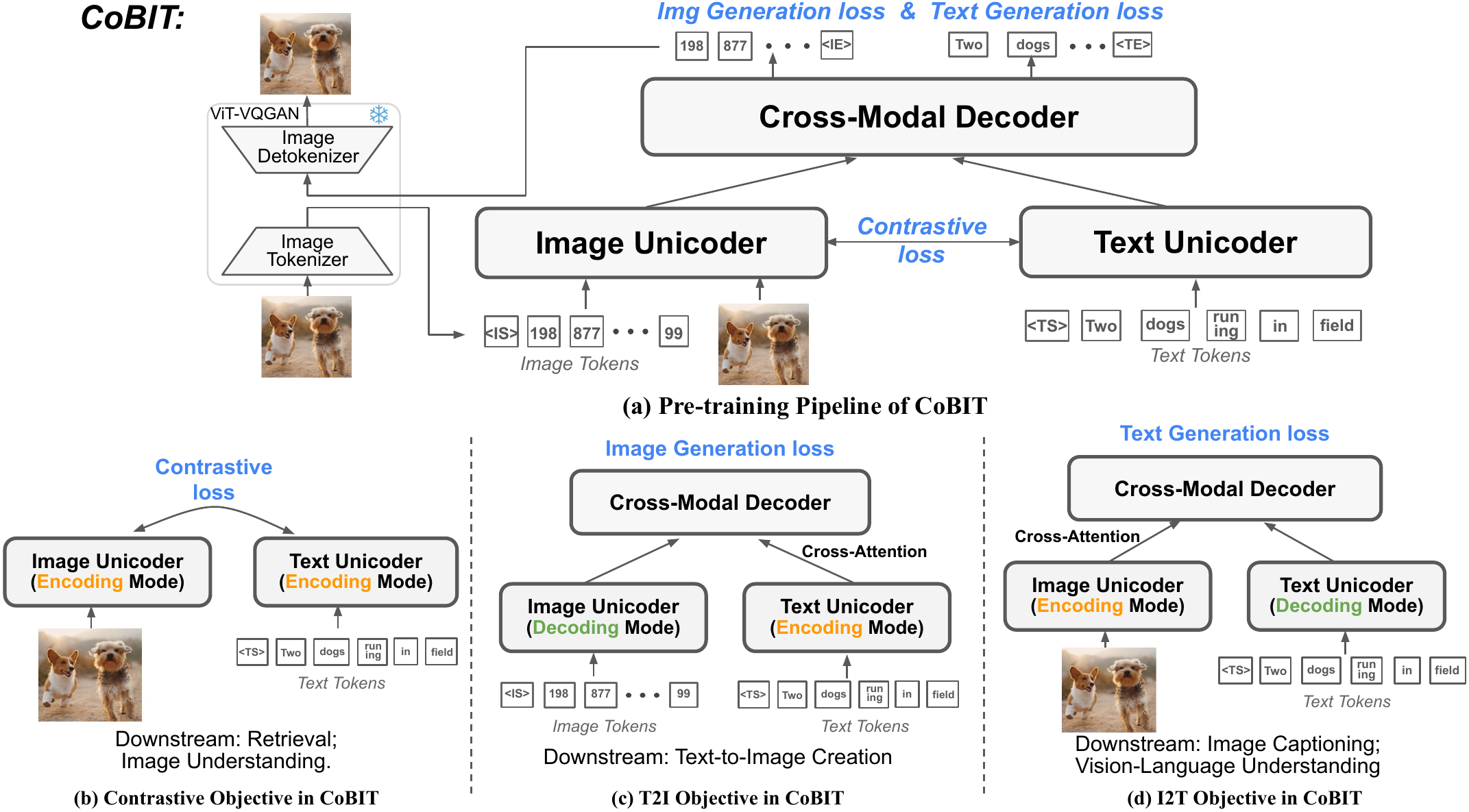}
  \vspace{-5mm}
  \caption{(a): Overview of \cocadraw{} pre-training pipeline; (b): When optimizing contrastive objective, image unicoder and text unicoder work as two encoders; (c) and (d): When optimizing image/text generation loss, text/image unicoder extracts features in encoding mode and image/text unicoder works in autoregressive decoding mode, then the cross-attention decoder will let autoregressive image/text features cross-attend to encoded text/image feature.
}
  \label{fig:diagram}
  \vspace{-5mm}
\end{figure*}

In this work, we aim to unify the three learning objectives: cross-modal contrastive learning, image-to-text generation and text-to-image generation, and thus consolidate the strengths of them in one framework.
We present a simple and unified \textbf{Co}ntrastive \textbf{B}i-directional \textbf{I}mage-\textbf{T}ext generation model (\bfcocadraw), which consists of an \textit{image unicoder} and a \textit{text unicoder}, as well as a cross-attention decoder. The proposed image/text unicoder structurally is Transformer, but they can switch in-between two modes: unimodal image/text encoding and decoding, which reuse the same set of Transformer parameters and only differ in input embeddings and attention masks. As shown in Fig. \ref{fig:diagram}, when optimizing contrastive objective, image unicoder and text unicoder work as two encoders. When optimizing text/image generation loss, image/text unicoder extracts features in encoding mode and text/image unicoder works in autoregressive decoding mode, then the cross-attention decoder will let autoregressive text/image features cross-attend to encoded image/text feature, serving as a fuser and generator. Each unicoder efficiently shares the knowledge between encoding and decoding, and therefore can jointly improve both T2I and I2T generation. In such a way, all three pre-training paradigms are unified in our framework.

\cocadraw{} is trained on both large-scale noisy web-crawled image-text data and image annotation data by treating labels as text. The pre-trained \cocadraw{}  subsumes strong zero-shot and transferable capacities of unimodal visual understanding, image-text matching, image-text understanding and text-to-image generation. For example, \cocadraw{} achieves 82.7\% accuracy in zero-shot ImageNet classification, 9.37 FID in zero-shot text-to-image generation, 44.8 CIDEr score in zero-shot image-to-text captioning. After fine-tuning, \cocadraw{} further achieves 86.44\% linear probing accuracy on ImageNet, 4.62 FID on text-to-image generation, and 78.3 VQA score.

\section{Related Work}
\noindent\textbf{Learning Visual Representation from Text.} Recent works studied pre-training a visual backbone with supervision from paired text data. CLIP \cite{radford2021learning} and ALIGN \cite{jia2021scaling} are two prominent large-scale models that used contrastive learning to pull together paired image-text embeddings while repelling unpaired ones. The resulting models exhibit strong visual zero-shot capacity as well as transferable visual representations. Florence \cite{yuan2021florence}, BASIC \cite{pham2021combined} and LiT \cite{zhai2022lit} further scale both datasets and models. FILIP \cite{yao2021filip} proposes to employ local token features from image and text for fine-grained contrastive learning. MS-CLIP \cite{you2022learning} and CLIPPO \cite{tschannen2022image} study sharing the model parameters between vision and text.

\noindent\textbf{Vision-Language Pre-training. } Another line of research focuses on learning a strong joint multimodal embedding of vision and language through pre-training. Some pre-train with mask-reconstruction loss \cite{li2019visualbert, wang2022image, li2022blip, chen2019uniter, shen2021much, li2021align}, \ie, mask partial image and text tokens in input and require the model to predict the masked tokens. Others pre-train models by generating text autoregressively  \cite{wang2021simvlm, chen2022pali, wang2022git,alayrac2022flamingo}. Both exhibit strong performance in downstream vision-language understanding tasks, such as VQA \cite{antol2015vqa} and captioning.

\noindent\textbf{Text-to-Image Generation.} Text-guided image creation is a challenging problem that has attracted intense interest in the past two years. Two families of methods are widely studied: diffusion-based and token-based. Diffusion-based models \cite{rombach2022high, saharia2022photorealistic, ramesh2022hierarchical} are based on a process that iteratively adds noise to images and then learns to reverse the noising process, while conditioning on textual descriptions of the image. With token-based methods, raw image are quantized into image tokens by an image tokenizer; then, given text input, Transformer models are used to predict image tokens autoregressively in a manner similar to machine translation \cite{ramesh2021zero, yu2022scaling} or by iteratively predicting image tokens in parallel\cite{chang2022maskgit, chang2023muse}.  

As these three broad lines of research have demonstrated great transferable ability to various downstream tasks, there have been many efforts to unify some of them \cite{yu2022coca, wang2022ofa, lu2022unified, zhang2021ernie, kim2022verse, huang2021unifying}. 
Our work, \cocadraw, serves as the first effort to integrate contrastive loss, image-to-text generation and text-to-image loss under one unified pre-training framework. 

\section{The \cocadraw{} Model}
We begin with describing input processing and then present the model architecture, which includes proposed unicoder module that shares the merit of both unimodal encoding and decoding. Finally, we provide a detailed explanation of the pre-training.

\subsection{Input}
To cover various tasks,
our model supports three types of input: text tokens, discrete image tokens, and raw images.

\noindent\textbf{Text Tokens.} Following the default process in past work ~\cite{raffel2020exploring, jia2021scaling, yu2022coca}, we tokenize text inputs using a SentencePiece model with a 64k vocabulary trained on the sampled pre-training datasets. The maximum text token length is 64. 

\noindent\textbf{Discrete Image Tokens.} \cocadraw{} generates images in an autoregressive manner, which requires tokenizing 2D images into a sequence of image tokens \cite{ramesh2021zero, ding2021cogview, ding2022cogview2, gafni2022make, yu2022scaling}. Following Parti~\cite{yu2022scaling}, we employ a pre-trained and frozen ViT-VQGAN \cite{yu2021vector} as the tokenizer. Specifically, each 256$\times$256 image is tokenized into a 32$\times$32 grid of image tokens, with 8192 image token classes in the codebook. We append the codebook to the text vocabulary as additional tokens. In inference, to generate images, we decode the image tokens one-by-one and feed them into the decoder in ViT-VQGAN to reconstruct the raw image. 
\label{sec:model_input}

\noindent\textbf{Raw Image.} For the purpose of image understanding and image-text understanding tasks, we also input raw images because it preserves the original information in pixels. Then each image is divided into non-overlapped patches following the de facto process in ViTs. In default, unless specified, the image resolution is 288x288 and the patch size is 18x18.

\subsection{Architecture}
\label{sec:arch}
 As shown in Fig. \ref{fig:diagram}, \cocadraw{} is composed of one image unicoder, one text unicoder and one cross-attention decoder. We term them unicoders because they can act as either encoders or decoders, depending on role they play for each task. The incorporation of text/image unicoder is inspired by ~\cite{dong2019unified, bao2020unilmv2, zhou2020unified}, which demonstrated that one Transformer model can perform both bidirectional encoding for understanding tasks and autoregressive decoding for generation tasks. In our scenario, compared with plain image/text encoders, unicoders in decoding mode can take advantage of the common knowledge shared with encoding to produce unimodal autoregressive features as a decent prior for cross-modal generative objective. Experimental ablation also validates that unicoders boost both T2I generation and multimodal understanding. 

\begin{table*}[t]
\small
\setlength\tabcolsep{10pt} %
\renewcommand{\arraystretch}{0.9} %
\centering
\begin{tabular}{lccccccc}
\toprule 
\multirow{2}{*}{Model} & \multicolumn{2}{c}{Image Unicoder} & \multicolumn{2}{c}{Text Unicoder} & \multicolumn{2}{c}{Cross-modal Decoder} &  \multirow{2}{*}{Total Params}\\
\cmidrule(lr){2-3} \cmidrule(lr){4-5} \cmidrule(lr){6-7}
 & Layers & Dims & Layers & Dims & Layers & Dims & \\
 \midrule
 \cocadrawbase & 12 & 768 & 12 & 768 & 18 & 1024 & 626M  \\
 \cocadrawlarge & 20 & 1024 & 12 & 1024 & 30 & 1024 & 1091M  \\
    \bottomrule
\end{tabular}
\vspace{-2mm}
\caption{Size variants of \cocadraw. }
\label{tab:model_scale}
\vspace{-6mm}
\end{table*}

\noindent\textbf{Image Unicoder.}  Recently, Vision Transformers (ViT) \cite{dosovitskiy2020image, touvron2021training, liu2021swin} has been established as the strongest approach for image feature 
\textit{encoding}. As \textit{decoders}, Transformers are used in autoregressive image token generation \cite{ramesh2021zero, gafni2022make, yu2022scaling}. We combine these two functionalities into a single image unicoder.  The image unicoder has two working modes: (1) In the encoding mode, following ViT, each 2D patch in the raw image is projected into  a feature vector by a trainable linear projection layer. Then, the sequence of projected features is input into cascaded Transformer layers to obtain the encoded image features, where the attention mask is bi-directional. (2) In the decoding mode, firstly, the input processing is different. As described in Sec. \ref{sec:model_input}, we tokenize the raw image into image tokens and initialize an embedding layer where token embeddings are indexed. Then, the same Transformer layers in encoding mode are reused in decoding mode to process the features; however, to guarantee the causal decoding ability, we use causal conv-shaped attention mask \cite{ramesh2021zero, yu2022scaling, child2019generating} instead. Overall, the two modes share the Transformer layers' parameters, and only differ in input processing and attention masks. Our assumption is that, compared with the design of plain image encoders as in previous work \cite{yu2022coca, wang2022git}, the additional decoding mode can exploit the common knowledge learned in image encoding to generate image autoregressive features, which we hypothesize should boost the (text-to-)image generation capacity.

\noindent\textbf{Text Unicoder.} Similar to image unicoder mentioned above, the text unicoder also has both encoding and decoding modes, which reuse the Transformer parameters. In both modes, the same tokenizer and embedding layer are utilized to obtain token features, given that they share the same input formats.
A causal attention mask is applied in decoding mode. During encoding of text, there are two options in previous work: bi-directional mask \cite{devlin2018bert, raffel2020exploring, yu2022scaling}, or causal mask \cite{brown2020language, radford2021learning, yao2021filip}. We empirically found that two masks make no difference in performance, and use causal masking as the default in the reported experiments.

\noindent\textbf{Cross-modal Decoder} The cross-modal decoder performs as a fusion-and-generation module, which structure-wise follows the cross-attention decoder \cite{vaswani2017attention, yu2022coca}. When generating text, the input is the text autoregressive feature from the text unicoder in decoding mode; encoded image features will be treated as cross-attention information, \ie, key and value in cross-attention layers. When generating the image, symmetrically, the image token autoregressive feature from the image unicoder in decoding mode is input and cross-attends to encoded text features. Also,  different from text generation where plain causal (autoregressive) mask is used in cross-modal decoder, image generation employs a conv-shaped masked sparse attention \cite{ramesh2021zero, yu2022scaling, child2019generating}, which can save memory and computation brought by long sequences of image tokens.

\subsection{Pre-training}
\label{sec:pretrain}
The pre-training of \cocadraw{} subsumes three fundamental objectives: image-text contrastive loss, I2T generation loss, T2I generation loss. Here, we provide details on the losses and also clarify the scaling and initialization strategy.

\noindent\textbf{Contrastive Loss.} We input raw image and text into the image unicoder and the text unicoder, respectively (both in encoding mode) to get encoded image and text features. For text, as with CLIP \cite{radford2019language} and ALIGN \cite{jia2021scaling}, we take the feature vector of the CLS token appended at the end of input sequence as the global representation. For images, however, the unicoder outputs a sequence of features. To aggregate them, following \cite{yu2022coca}, we apply an attention pooler, which is a single multi-head
attention layer with one learnable query and unicoder output features as key and value. After obtaining two global features of image and text, a contrastive loss is applied to optimize the paired image-text against others in the same batch:

\begin{footnotesize}
\begin{equation}
    \mathcal{L}_{\text{Con}}=- \frac{1}{N} (\sum_{i}^{N} \text{log} \frac{\text{exp}(x_i^T y_i / \tau)}{\sum_{j=1}^{N} \text{exp}(x_i^T y_j/ \tau)} +  \sum_{i}^{N} \text{log} \frac{\text{exp}(y_i^T x_i / \tau)}{\sum_{j=1}^{N} \text{exp}(y_i^T x_j/ \tau)}),
\end{equation}
\end{footnotesize}
where $x_i$ and $y_j$ denote the normalized global embeddings of $i$-th image and $j$-th text. $\tau$ is a learnable temperature for adjusting the scale of the loss.

\begin{table*}[t]
\small
\setlength\tabcolsep{4pt} %
\renewcommand{\arraystretch}{1.0} %
\centering
\begin{tabular}{lccccccccccc}
\toprule
\multirow{3}{*}{Model} & Image Understd. & \multicolumn{9}{c}{Image-Text Understd.} & Content Creation \\
\cmidrule(lr){2-2} \cmidrule(lr){3-11} \cmidrule(lr){12-12}
&\multirow{2}{*}{\makecell[c]{ImageNet \\ Classification}} &\multicolumn{4}{c}{Flickr Retrieval} & \multicolumn{4}{c}{MS-COCO Retrieval} & \multirow{2}{*}{\makecell[c]{MS-COCO \\ Captioning}}   & \multirow{2}{*}{\makecell[c]{MS-COCO \\ T2I Generation}}\\
\cmidrule(lr){3-6} \cmidrule(lr){7-10}
&  &\multicolumn{2}{c}{Image$\rightarrow$Text}&\multicolumn{2}{c}{Text$\rightarrow$Image}&\multicolumn{2}{c}{Image$\rightarrow$Text}&\multicolumn{2}{c}{Text$\rightarrow$Image}& &  \\
\cmidrule(lr){2-2} \cmidrule(lr){3-4} \cmidrule(lr){5-6} \cmidrule(lr){7-8}  \cmidrule(lr){9-10} \cmidrule(lr){11-11} \cmidrule(lr){12-12}
& Acc(\%) & R@1 &R@5 &R@1 &R@5 & R@1 &R@5 &R@1 &R@5 & CIDEr & FID ($\downarrow$) \\
\midrule
CLIP \cite{radford2021learning} & 76.2 & 88.0 & 98.7 & 68.7 & 90.6 & 58.4 & 81.5 & 37.8 & 62.4 & - & - \\
ALIGN \cite{jia2021scaling} & 76.4  &  88.6 & 98.7 & 75.7 & 93.8 & 58.6 & 83.0 & 45.6 & 69.8 & - & - \\
FILIP \cite{yao2021filip} & 78.3 & 89.8 &  99.2 & 75.0 &  93.4  & 61.3 &  84.3  & 45.9 &  70.6 & - & - \\
Florence \cite{yuan2021florence} & \red{83.7} & 90.9 & 99.1 & 76.7 & 93.6 & 64.7 & \redbf{85.9} & 47.2 & 71.4 & - & - \\
CoCa-Large \cite{yu2022coca} & \redbf{84.8} & \red{91.4} & \redbf{99.2} &  \red{79.0} & \red{95.1} & \redbf{65.4} & \red{85.6} & \red{50.1} & \red{73.8} & - & - \\
ZeroCap \cite{tewel2021zero} & -  & - & - & - & - & - & - & - & - & 14.6 & - \\
SimVLM \cite{wang2021simvlm} & -  & - & - & - & - & - & - & - & - & 32.2 & - \\
VLKD \cite{dai2022enabling} & -  & - & - & - & - & - & - & - & - & \lb{58.3\textsuperscript{\textdagger}} & - \\
Parti-350M \cite{yu2022scaling} & -  & - & - & - & - & - & - & - & - & - & 14.10 \\ 
Parti-750M \cite{yu2022scaling} & -  & - & - & - & - & - & - & - & - & - & 10.71 \\ 
LDM-1.4B \cite{rombach2022high} & -  & - & - & - & - & - & - & - & - & - & 12.63 \\
\midrule
\rowcolor{lg} 
\lb{Coca-2B \cite{yu2022coca}} & \lb{86.3} & \lb{92.5} & \lb{99.5} & \lb{80.4} & \lb{95.7} & \lb{66.3} & \lb{86.2} & \lb{51.2} & \lb{74.2} & - & - \\ 
\rowcolor{lg}
\lb{Flamingo-3B \cite{alayrac2022flamingo}} & -  & - & - & - & - & - & - & - & - & \lb{73.0\textsuperscript{\textdagger}}  & -  \\
\rowcolor{lg}
\lb{DALL-E 2 \cite{ramesh2022hierarchical}} & -  & - & - & - & - & - & - & - & - & - & \lb{10.39} \\
\rowcolor{lg}
\lb{Parti-20B \cite{yu2022scaling}} & -  & - & - & - & - & - & - & - & - & - & \lb{7.23} \\
\midrule
\cocadrawbase{} & 79.4 & 89.5 & 98.4 & 76.5 & 94.3 & 62.1 & 83.5 & 47.3 & 72.3 & \red{43.0} & \red{10.35} \\
\cocadrawlarge{} & 82.7 & \redbf{91.5} & \red{99.1} & \redbf{79.9} & \redbf{95.3} & \red{65.1} & 85.5 & \redbf{50.3} & \redbf{74.2} & \redbf{44.8} & \redbf{9.37} \\
\bottomrule
\end{tabular}
\vspace{-2mm}
\caption{\textbf{Zeroshot Evaluation} of \cocadraw{} against previous image-text models.  The models in \colorbox{lg}{gray} background have $\gg$ 1B parameters while others in white background have $\lesssim$ 1B parameters. For models with $\lesssim$ 1B parameters , we highlight the best score in \redbf{bold+underline} and the second-best score in \red{underline}. Understd. is the abbreviation of Understanding. }
\label{tab:exp_zs}
\vspace{-5mm}
\end{table*}

\noindent\textbf{I2T and T2I Generation Loss. } We formulate two generation tasks as token generation problems. As shown in Fig. \ref{fig:diagram}, by cascading image unicoder, text unicoder and cross-modal decoder, we can perform two tasks seamlessly by only switching the working modes of unicoders. A cross-entropy loss is applied on top of cross-modal decoder to maximize the conditional likelihood of the ground-truth token under the forward autoregressive factorization.
\begin{equation}
    \mathcal{L}_{\text{I2T}} = -\sum_{t=1}^{T} \text{log} P_\theta (y_t|y_{<t}, I), \\
\end{equation}
\begin{equation}
    \mathcal{L}_{\text{T2I}} = -\sum_{t=1}^{T} \text{log} P_\theta (x_t|x_{<t}, T), \\
\end{equation}
where $y$ and $x$ denote text and image tokens respectively. 

\noindent\textbf{Classifier-Free Guidance for T2I. } Following \cite{yu2022scaling, chang2023muse, ramesh2021zero}, we employ classifier-free guidance (CFG) \cite{ho2022classifier} in text-to-image generation. To be more specific, in training, we randomly mask conditioning vectors, \ie, input text tokens, by certain possibility (10\% in our implementation). In inference, we compute two predictions: conditional one $I(z, T)$ and unconditional one $I(z)$, which only differ in text input: conditional prediction  $I(z, T)$ has original text tokens as input while the input text of unconditional prediction $I(z)$ is fully masked. Then we linearly interpolate the $I(z,c)$ and $I(z)$ to obtain the final generated image:
\begin{equation}
    I = I(z) + \alpha (I(z,T) - I(z)),
\end{equation}
where $\alpha$ is a hyperparameter for adjusting the scale of classifier-free guidance, and we set $\alpha$=2.0 in default.

\noindent\textbf{Final Loss. } In the end, we simply add those three losses up to optimize the model end-to-end.
\begin{equation}
\mathcal{L}_{\text{\cocadraw}} = \lambda_{\text{Con}} \mathcal{L}_{\text{Con}} + \lambda_{\text{I2T}} \mathcal{L}_{\text{I2T}} + \lambda_{\text{T2I}} \mathcal{L}_{\text{T2I}}
\end{equation}
where $\lambda_{\text{Con}}$, $ \lambda_{\text{I2T}}$, $ \lambda_{\text{T2I}}$ denote corresponding scalar coefficients for contrastive, I2T and T2I loss. In default, we set $\lambda_{\text{T2I}}$ : $ \lambda_{\text{I2T}}$ : $ \lambda_{\text{Con}}$ = 1 : 0.2 : 0.1. 

\noindent\textbf{Scaling.}  As shown in Tab.\ref{tab:model_scale}, we start from \cocadrawbase{}, and scale it up, w.r.t. both number of layers and model dimension, to obtain \cocadrawlarge{} with around 1B parameters. 

\noindent\textbf{Initialization.}  In previous text-to-image generation models \cite{yu2022scaling, rombach2022high, saharia2022photorealistic}, the text feature extractor is usually initialized by a pre-trained text model. Correspondingly, in \cocadraw{}, we also initialize text unicoder with pre-trained text unimodal decoder from CoCa \cite{yu2022coca}, while leaving the image unicoder and cross-modal decoder trained from scratch. In Sec. \ref{sec:exp_ablation}, we compare it with training all from scratch, and find the initialization indeed helps with a small margin.  


\section{Experiments}
In this section, we first describe the pre-training data and optimization process (Sec. \ref{sec:exp_setup}). Sec. \ref{sec:exp_zs} and Sec. \ref{sec:exp_ft} detail the primary results of zero-shot evaluation and fine-tuning evaluation, respectively.  Both evaluations examine three capacities: (1) visual understanding, (2) image captioning and multimodal understanding, (3) text-to-image content creation. In Sec. \ref{sec:exp_ablation}, different components of \cocadraw{} are ablated to justify our design.

\begin{table*}[t]
\small
\setlength\tabcolsep{4pt} %
\centering
\begin{tabular}{lcccccccc}
\toprule
\multirow{3}{*}{Model} & \multirow{3}{*}{\makecell[c]{Visual \\ Backbone}} & Image Understd. & \multicolumn{5}{c}{Image-Text Understd.} & Content Creation \\
\cmidrule(lr){3-3} \cmidrule(lr){4-8} \cmidrule(lr){9-9}
& & \multirow{2}{*}{\makecell[c]{ImageNet  \\ Linear Probing }} &\multicolumn{2}{c}{VQA} & \multicolumn{2}{c}{SNLI-VE} & \multirow{2}{*}{\makecell[c]{MS-COCO \\ Captioning \scriptsize{(CIDEr)}}}   & \multirow{2}{*}{\makecell[c]{MS-COCO T2I \\  Generation \scriptsize{(FID$\downarrow$)}}}\\
\cmidrule(lr){4-5} \cmidrule(lr){6-7}
&  & &test-dev&test-std&dev&test& &  \\
\midrule
CLIP \cite{radford2021learning}  & Scratch & 85.4 &  -   & -  & -   & -  & - & - \\
ALIGN \cite{jia2021scaling}  & Scratch & \red{85.5} &  -   & -  & -   & -  & - & - \\
UNITER \cite{chen2019uniter}  & Faster-RCNN & - &  73.8  & 74.0 & 79.4  & 79.4 & - & - \\
VinVL \cite{zhang2021vinvl} & Faster-RCNN & - &  76.5 &76.6 & - & - & 130.8 & - \\
CLIP-ViL \cite{shen2021much} & CLIP & - &  76.5   &  76.7 & 80.6 & 80.2 & 134.2 & - \\
ALBEF \cite{li2021align} & PT. ViT & - &  75.8  & 76.0 & 80.8  & 80.9  & - & - \\
BLIP \cite{li2022blip} & PT. ViT & - &  78.3  & 78.3  & -   & -  & 136.7 & - \\
SimVLM \cite{wang2021simvlm} & PT. ResNet  & - &   \red{80.0} & \red{80.3}  & \redbf{86.2} & \redbf{86.3} &  \red{143.3} & - \\
OFA \cite{wang2022ofa} & PT. ResNet  & - 
&   \redbf{82.0} & \redbf{82.0}  & \llb{91.0} \textsuperscript{\textdagger}  & \llb{91.2}\textsuperscript{\textdagger} &  \redbf{145.3} & 10.5 \\
X-LXMERT \cite{huang2021unifying} & Faster-RCNN  & - &   - & -  & - & - &  122.6 &  29.9 \\
\midrule
\rowcolor{lg} 
\lb{CoCa-2.1B \cite{yu2022coca}} &  \lb{Scratch}  & - &   \lb{80.0} & \lb{80.3}  & \lb{87.0} & \lb{87.1} &  \lb{143.3} & - \\
\rowcolor{lg} 
\lb{BEIT3-1.9B \cite{wang2022image}} & \lb{Scratch}  & - 
&   \lb{84.2} & \lb{84.0}  & -  & - &  \lb{147.6} & - \\
\rowcolor{lg} 
\lb{PALI-17B \cite{chen2022pali}} & \lb{PT. ViT}  & - 
&   \lb{84.3}  & \lb{84.3}  & -  & - &  \lb{149.1} & - \\
\rowcolor{lg}
\lb{Parti-20B \cite{yu2022scaling}} & -  & - & - & - & - & - & - & \lb{3.22} \\
\midrule
\cocadrawbase{} & Scratch  & 83.48  & 76.3 & 76.6  & 85.4  & 85.4 &  135.4 & \red{5.06} \\
\cocadrawlarge{} & Scratch  & \redbf{86.44}  & 77.9 & 78.3  & \redbf{86.2}  & \red{86.0} &  139.5 & \redbf{4.62} \\
\bottomrule
\end{tabular}
\vspace{-2mm}
\caption{\textbf{Fine-tuning Evaluation} of \cocadraw{} against previous image-text models. PT. denotes pre-trained and Scratch denotes trained from scratch. {}\textsuperscript{\textdagger}OFA incorporates both images and text in its input while others only use image one. }
\label{tab:exp_ft}
\vspace{-5mm}
\end{table*}

\subsection{Pre-training Details}
\label{sec:exp_setup}
\noindent\textbf{Data.} \cocadraw{} is designed to be pre-trained with image-text data.  For contrastive loss and I2T loss, we use a mixture of ALIGN dataset \cite{jia2021scaling}, and JFT-4B dataset \cite{zhai2022scaling} where category names are transformed into texts by prompts as in \cite{pham2021combined}. Differently, for T2I generation, we found that the short text in JFT-4B is less informative for generating the image as extensive descriptions of visual details are important. Instead, we replace JFT with WebLI dataset \cite{chen2022pali}, and mix it with ALIGN for T2I generation loss. 
We further perform de-duplication, as in \cite{jia2021scaling, zhai2022lit}, to remove the examples close to downstream tasks. In the end, we obtain 1.1B pairs from ALIGN dataset, 162M pairs from WebLI dataset and 4B pairs from JFT-4B dataset.

\noindent\textbf{Optimization. } Our \cocadraw{} models are implemented using Pax \cite{Pax}, a Jax-based framework. Within each batch, for optimizing T2I loss, we sample 1,024 image-text pairs from a mixture of ALIGN and WebLI datasets, and for optimizing contrastive and I2T losses, we sample 30,720 image-text pairs from a mixture of ALIGN and JFT datasets. In total, the batch size is 31,744. We use the Adafactor \cite{shazeer2018adafactor} optimizer with $\beta_1$ = 0.9, $\beta_2$ = 0.96 and a weight decay of 0.045. As for the learning rate schedule, we warm it up to 4.5e-5 in first 5,000 steps and then use an exponential decay starting from the step of 85,000. In total, models are pre-trained for 1M steps and \cocadrawbase/\cocadrawlarge{} takes around 12 days on 256/512 CloudTPUv4 chips. Then, following \cite{radford2021learning, jia2021scaling, yuan2021florence, yu2022coca}, we further pre-train our models for 50k steps with 576x576 high-resolution raw images as input in image encoding. The image input to ViT-VQGAN, \ie, image for decoding, is kept at 256x256 resolution.

\subsection{Zero-shot Evaluation on Downstream Tasks}
\label{sec:exp_zs}
\cocadraw{} is capable of versatile zero-shot abilities. By evaluating on 5 representative tasks, \cocadraw{} stands out as the first work able to perform image understanding, image-text understanding and text-guided content creation in the zero-shot manner, and achieves superior performance on the majority of evaluation metrics. 

\noindent\textbf{Zero-shot Image Classification. }  Following \cite{radford2021learning, jia2021scaling, yuan2021florence}, we apply the same set of prompts to transfer labels into sentences, such as ``a photo of \{class\}''. The same as the way of computing contrastive loss in Sec. \ref{sec:pretrain}, we input raw image/text into image/text unicoders in encoding mode to  obtain the global features of image and text.  Then we compute their similarity to match images and labels. As shown in Tab. \ref{tab:exp_zs}, compared to models with similar scales, in ImageNet \cite{russakovsky2015imagenet}, \cocadrawlarge{} can achieve 82.7 \%, outperforming strong baselines such as CLIP \cite{radford2021learning}, ALIGN \cite{jia2021scaling}. We can see there is still a 2\% gap between \cocadrawlarge{} and CoCa-Large \cite{yu2022coca}, which may come from batch size difference: CoCa's batch size is 64k while ours is 30k only.

\noindent\textbf{Zero-shot Image-Text Retrieval. } The image and text feature extraction process is the same as zero-shot image classification. Flick \cite{plummer2015flickr30k} and MS-COCO \cite{lin2014microsoft} are used for evaluation. In Tab. \ref{tab:exp_zs}, within comparable scales, \cocadrawlarge{} can outperform the previous best model CoCa-Large in 5 out of 8 metrics and is ranked the second best in another two metrics, which shows a superior vision-language understanding capacity.

\noindent\textbf{Zero-shot Image Captioning. } Since \cocadraw{} is already pre-trained with image-to-text generation loss on noisy image-text data, it's natural to directly evaluate it on zero-shot image captioning in the same way. As shown in Tab. \ref{tab:exp_zs}, in MS-COCO, \cocadrawbase/\cocadrawlarge{} can achieve 43.0/44.8 CIDEr score, surpassing SimVLM by 10.8/12.6. It's noted that the models with \textsuperscript{\textdagger}, \eg, Flamingo, VLKD, have much higher scores than others, because they reuse a pre-trained large language model as a decoder that inherits strong text generation ability.

\noindent\textbf{Zero-shot Text-to-Image Generation. } Similar to computing text-to-image generation in pre-training, we can also evaluate \cocadraw{} in zero-shot text-to-image generation. In decoding, following \cite{yu2022scaling, ramesh2021zero}, we employ Top-K sampling to sample 16 images for each text and use a reranker to select the best image for evaluation. Following the de facto process, we compute FID score \cite{heusel2017gans} on MS-COCO 30k data \cite{ramesh2021zero, saharia2022photorealistic} (lower FID is better). As we can see in Tab.\ref{tab:exp_zs}, \cocadraw s can beat specialized models with comparable scales, and \cocadrawlarge{} can achieve an impressive FID of 9.37 which outperforms some models with larger scale by a significant margin, \eg, DALL-E 2 with 3.5B parameters, Make-A-Scene (FID=11.84) \cite{gafni2022make} with 4B parameters. 

\begin{table}[t]
\centering
\renewcommand{\arraystretch}{0.9} %
\begin{tabular}{cccccc}
\toprule
 \multicolumn{3}{c}{Objectives} & \multicolumn{3}{c}{Evaluation} \\
 \cmidrule(lr){1-3}  \cmidrule(lr){4-6}
 Con. & T2I & I2T & ZS IN. & VQA. & ZS IG. ($\downarrow$) \\
 \midrule
 $\checkmark$ & - & - & 70.8 & - & -\\
 - &  $\checkmark$& - & - &  - & 12.6 \\
 - & - & $\checkmark$ & - & 68 & -\\
 - & $\checkmark$ & $\checkmark$ & - & 65.4 & 13.2\\
 $\checkmark$ & $\checkmark$ & $\checkmark$ & 71.1 & 66.9 & 13.3\\
\bottomrule
\end{tabular}
\vspace{-2mm}
\caption{Ablation on three objectives. Con. denotes contrastive loss.}
\label{tab:exp_ablate_obj}
\vspace{-2mm}
\end{table}

\subsection{Fine-tuning on Downstream Tasks}
\label{sec:exp_ft}
To demonstrate the transferability of \cocadraw{} on image understanding, image-text understanding and text-guided content creation, we further conduct linear probing or fine-tuning on multiple downstream tasks.

\noindent\textbf{Linear Probing on ImageNet. } Following \cite{radford2021learning, jia2021scaling}, we linear probe \cocadraw{} by fixing all parameters of image unicoder and only training a linear classifier on top for image recognition. \cocadrawlarge{} can outperform CLIP and ALIGN by around 1\%.

\noindent\textbf{Image-Text Understanding. } We categorize VQA \cite{antol2015vqa}, SNLI-VE \cite{xie2019visual} and image captioning into tasks requires image-text understanding. We fine-tune all parameters of \cocadraw{} and evaluate it on val/test set. 

\begin{table}[t]
\centering
\renewcommand{\arraystretch}{0.9} %
\begin{tabular}{ccccc}
\toprule
 \multicolumn{2}{c}{Module} & \multicolumn{3}{c}{Evaluation} \\
 \cmidrule(lr){1-2}  \cmidrule(lr){3-5}
 Image & Text & VQA & ZS Cap. & ZS IG. ($\downarrow$) \\
\midrule
Encoder & Encoder & 65.9 & 32.9  &  13.8\\
Unicoder & Encoder & 66.5 & 36.9 &  13.38\\
Encoder & Unicoder & 67.8 & 35.0  &  13.67\\
Unicoder & Unicoder & 66.9 & 37.9  & 13.31\\
\bottomrule
\end{tabular}
\vspace{-3mm}
\caption{Ablation on unicoder vs. encoder.}
\label{tab:exp_ablate_unicoder}
\vspace{-4mm}
\end{table}

\begin{table}[t]
\centering
\begin{tabular}{cccc}
\toprule
 \multirow{2}{*}{Model} & \multicolumn{3}{c}{Evaluation} \\
 \cmidrule(lr){2-4} 
 & ZS IN. & VQA & ZS IG. ($\downarrow$) \\
 \midrule
 \multirow{2}{*}{\makecell[c]{Init. Text Unicoder\\ from CoCa}} &  \multirow{2}{*}{75.35} &  \multirow{2}{*}{68.48} &  \multirow{2}{*}{11.42} \\
 &  &  &  \\ 
 \hline
 Train from Scratch &  75.02 & 68.55 & 11.63\\
\bottomrule
\end{tabular}
\vspace{-2mm}
\caption{Ablation on initialization. }
\label{tab:exp_ablate_init}
\vspace{-5mm}
\end{table}

\noindent\textbf{\textit{Captioning. }} In fine-tuning, \cocadraw{} computes caption predictions in the same way as zero-shot image captioning in Sec. \ref{sec:exp_zs}. In Tab. \ref{tab:exp_ft}, we can see the \cocadraw{} can achieve a competitive CIDEr score against other models. It's noted that some works \cite{wang2022ofa} additionally apply task-specific tricks such as CIDEr optimization, but for a fair comparison, we only present their results with plain cross-entropy loss . 

\noindent\textbf{\textit{VQA. }}   Following prior works \cite{wang2021simvlm, yu2022coca}, we use the VQA v2 and the task is formulated as a classification problem over 3,129 most frequent answers in the training set. To accomplish this, the raw image is fed into the image unicoder using encoding mode, while the question is processed by the text unicoder in decoding mode. Subsequently, the cross-modal decoder utilizes the text decoding features as input and cross-attends to the encoded image features. The final token output feature of the cross-modal decoder is considered the fused global feature. To predict the answer, a linear classifier is trained on top of this feature. As shown in Tab. \ref{tab:exp_ft}, \cocadraw{} can achieve satisfactory performance compared with other VLP models.

\noindent\textbf{\textit{SNLI-VE. }} Similar to fine-tuning VQA, we extract the final token output feature of cross-modal decoder and apply a linear classifier on top to predict the three relations. As shown in Tab. \ref{tab:exp_ft}, \cocadraw{} can outperform strong VLP models and achieve superior performance. Note that other models including \cocadraw{} only use image premise as inputs, but OFA incorporates both image and text premises in its input.

\begin{figure*}[t]
  \includegraphics[width=\linewidth]{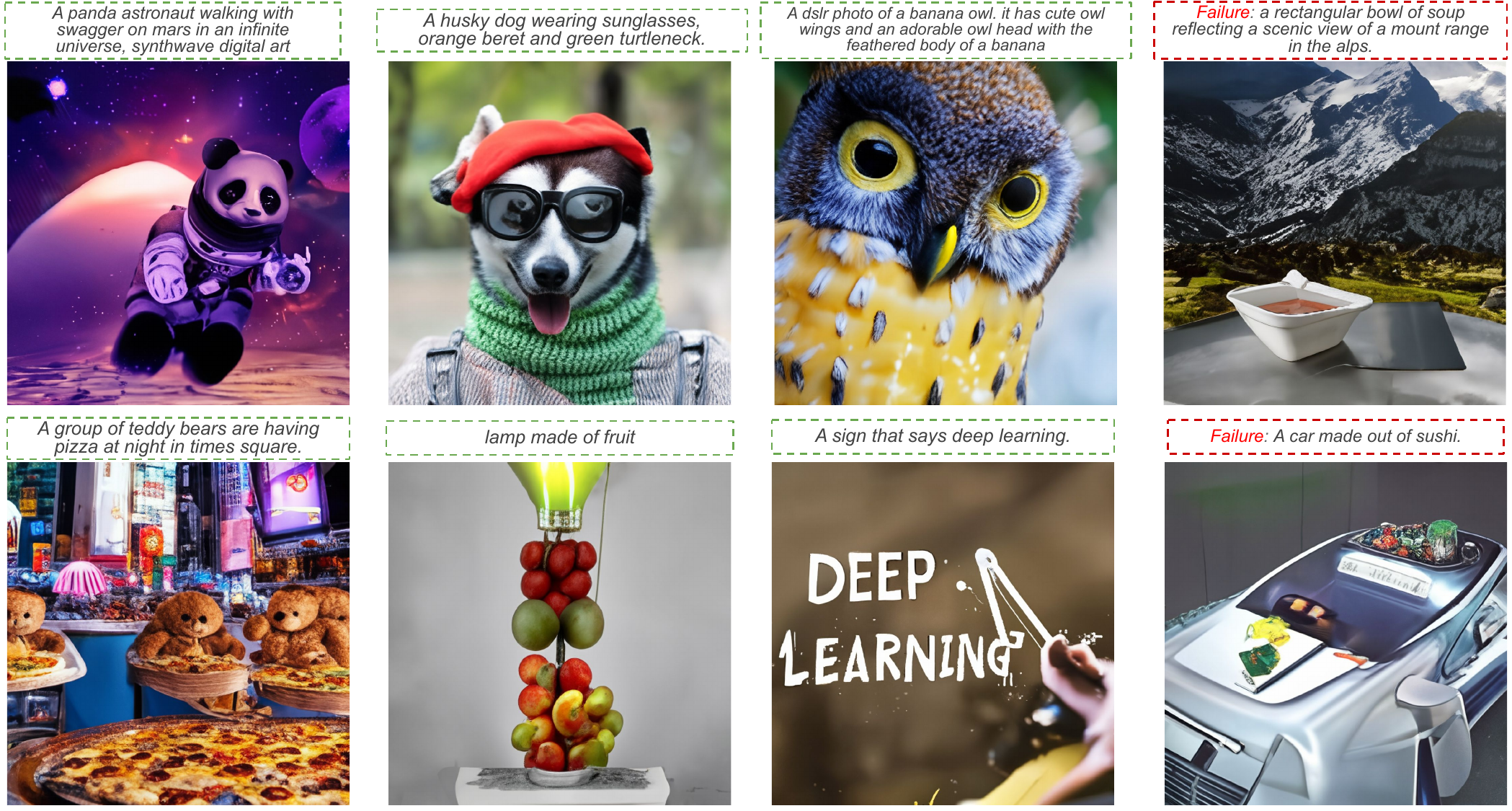}
  \vspace{-6mm}
  \caption{Qualitative results of zero-shot text-to-image generation from \cocadrawlarge{} with both good and failed cases. }
  \label{fig:vis}
\vspace{-5mm}
\end{figure*}

\noindent\textbf{Text-to-Image Generation. } Following \cite{yu2022scaling, saharia2022photorealistic}, we fine-tune \cocadraw{} on MS-COCO training set and evaluate the FID score on sampled 30k test set. Compared with zero-shot performance, fine-tuning on \cocadrawbase/\cocadrawlarge{} further reduces the FID from 10.35/9.37 to 5.06/4.62, outperforming models of comparable scales.

\subsection{Ablation}
\label{sec:exp_ablation}
In this section, we comprehensively ablate the design choices in \cocadraw{}. Most ablation experiments are conducted on \cocadrawbase{} with a reduced batch size and a shrunken training schedule. Specifically, total batch size is 4,352, containing 4,096 for contrastive and I2T loss, and 256 for T2I loss, and the total training step is 200k, without high-resolution pre-training.  We select following representative tasks: zero-shot ImageNet Classification (ZS IN.) for image understanding, fine-tuned VQA or MS-COCO zero-shot Captioning (ZS Cap.) for multimodal understanding, MS-COCO zero-shot text-to-image generation (ZS IG.) for image generation. ZS Cap. result is measured by CIDEr and ZS IG. result is measured in FID (lower FID is better).

\noindent\textbf{Training Objectives. } We ablate the existence of three training objectives: contrastive loss, I2T loss and T2I loss, and study how they affect each other. The result is shown in Tab. \ref{tab:exp_ablate_obj}. We can obtain several interesting observations: (1). By comparing the first and last rows, it's found \textit{cross-modal generation objectives can improve image understanding a bit on top of contrastive loss}. The zero-shot ImageNet accuracy is improved by 0.3\%. (2) Comparing the second, third and fourth rows, we see \textit{two generations losses, \ie, I2T loss and T2I loss, contradict each other a little bit}. After adding T2I loss, VQA accuracy drops by 2.6, and after adding I2T loss, zero-shot image generation FID score rises by 0.6. (3) From the fourth row and fifth row, we can see \textit{contrastive loss improves vision-language understanding while it doesn't influence image generation}. Overall, we demonstrate the feasibility of unifying three fundamental objectives in one framework relatively harmoniously.

\noindent\textbf{Loss Weight. } Given three objectives, we ablate different weights for them and select the best one as the default configuration for all experiments. Please see supplementary.

\noindent\textbf{Unicoder vs. Encoder. } In previous Vision-Language works \cite{wang2021simvlm, yu2022coca, chen2022pali}, encoder-decoder has been a de facto pipeline, where encoder encodes image/text features and cross-modal decoder fuses them and perform generation. Differently, we propose unicoder to replace encoder, which can both encode and decode unimodal representations with shared parameters. Here, we ablate image and text unicoders against image and text encoders. It's noted that unicoder doesn't add extra parameters to encoders because encoding and decoding in unicoder reuse same set of parameters. We put a diagram in supplementary to illustrate how the compared encoder-only models work.  As shown in Tab. \ref{tab:exp_ablate_unicoder}, either image unicoder or text unicoder can improve over encoders, and applying them together brings best trade-off for both image generation and multimodal understanding.  

\noindent\textbf{Train From Scratch. } As mentioned in Sec. \ref{sec:pretrain}, we initialize text unicoder with a pre-trained unimodal text decoder from CoCa. Here we also attempt to train all from scratch. In this comparison, all models are trained with non-shrunken batch size to mitigate the possible gap due to much larger batch size of CoCa. In Tab. \ref{tab:exp_ablate_init}, loading pre-trained weight from CoCa improves zero-shot Imagenet recognition and text-to-image generation by 0.3\% and 0.2, which is a small margin. Also, it doesn't even improve VQA. This comparison verifies the do-ability of training \cocadraw{} all from scratch without hurting much performance.

\subsection{Visualization}
We visualized good and failed generated images of \cocadrawlarge{} using the prompts from PartiPrompt \cite{yu2022scaling}. As in Fig. \ref{fig:vis}, \cocadraw{} can generate high-quality, broadly capable, open-domain images based on text. As for failed cases, we can see \cocadraw{} misunderstands ``A car made out of Sushi'' as ``A car with Sushi on top'', also \cocadraw{} fails to generate the reflection of mountains in the bowl of soup. More visualization and analysis are in supplementary.


\section{Conclusion}
We present a vision-language foundation model, CoBIT, which unifies three objectives: cross-modal contrastive learning, image-to-text generation, and text-to-image generation. CoBIT consists of an image unicoder, a text unicoder, and a cross-attention decoder. The unicoders can switch between two modes: unimodal image/text encoding and decoding. The model is trained on large-scale noisy web-crawled image-text data and image annotation data. CoBIT achieves strong zero-shot and transferable capacities of unimodal visual understanding, image-text matching, image-text understanding, and text-to-image content creation. 

\vspace{-2mm}
\section*{Limitations \& Broader Impact}
\vspace{-1mm}
\noindent\textbf{Limitations. } Although \cocadraw{} unifies contrastive loss, text-to-image generation (T2I) loss, and image-to-text (I2T) generation loss, from ablation experiments, we nevertheless find that the T2I and I2T objectives contradict each other somewhat. We hypothesize that it is because the two different types of generation require fine-grained knowledge that is specific to each modality.

\noindent\textbf{Broader Impact. } Models such as Stable Diffusion, DALL-E, Parti, and \cocadraw{} are trained on large and noisy  image-text datasets that include many biases due to the distribution of images contained in them (which is not representative of real world, generally speaking) and the way the images are described (which includes human biases in descriptions of different subjects). Such models also create risks with respect to misinformation (e.g. deepfakes) and they introduce both challenges and opportunities for creativity and art. See the Parti \cite{yu2022scaling} and Imagen \cite{saharia2022photorealistic} broader impacts sections for extensive discussion and references on these topics. Furthermore, the task setting of creating images from descriptions itself involves inherent ambiguities (especially due to underspecification) that play an important role -- beyond the training data -- in the behavior and risks inherent in such models \cite{hutchinson-etal-2022-underspecification}.

\vspace{-3mm}
\section*{Acknowledgement}
\vspace{-2mm}
We would like to thank Prof. Shih-Fu Chang, Luowei Zhou, Long Chen for constructive discussion, Zirui Wang for help with downstream fine-tuning, Huiwen Chang for proofreading, and Laurent El Shafey for infra support.





{\small
\bibliographystyle{ieee_fullname}
\bibliography{egbib}

\begin{thebibliography}{10}\itemsep=-1pt

\bibitem{alayrac2022flamingo}
Jean-Baptiste Alayrac, Jeff Donahue, Pauline Luc, Antoine Miech, Iain Barr,
  Yana Hasson, Karel Lenc, Arthur Mensch, Katie Millican, Malcolm Reynolds,
  et~al.
\newblock Flamingo: a visual language model for few-shot learning.
\newblock {\em arXiv preprint arXiv:2204.14198}, 2022.

\bibitem{antol2015vqa}
Stanislaw Antol, Aishwarya Agrawal, Jiasen Lu, Margaret Mitchell, Dhruv Batra,
  C~Lawrence Zitnick, and Devi Parikh.
\newblock Vqa: Visual question answering.
\newblock In {\em Proceedings of the IEEE international conference on computer
  vision}, pages 2425--2433, 2015.

\bibitem{bao2020unilmv2}
Hangbo Bao, Li Dong, Furu Wei, Wenhui Wang, Nan Yang, Xiaodong Liu, Yu Wang,
  Jianfeng Gao, Songhao Piao, Ming Zhou, et~al.
\newblock Unilmv2: Pseudo-masked language models for unified language model
  pre-training.
\newblock In {\em International conference on machine learning}, pages
  642--652. PMLR, 2020.

\bibitem{brown2020language}
Tom Brown, Benjamin Mann, Nick Ryder, Melanie Subbiah, Jared~D Kaplan, Prafulla
  Dhariwal, Arvind Neelakantan, Pranav Shyam, Girish Sastry, Amanda Askell,
  et~al.
\newblock Language models are few-shot learners.
\newblock {\em Advances in neural information processing systems},
  33:1877--1901, 2020.

\bibitem{chang2023muse}
Huiwen Chang, Han Zhang, Jarred Barber, AJ Maschinot, Jose Lezama, Lu Jiang,
  Ming-Hsuan Yang, Kevin Murphy, William~T Freeman, Michael Rubinstein, et~al.
\newblock Muse: Text-to-image generation via masked generative transformers.
\newblock {\em arXiv preprint arXiv:2301.00704}, 2023.

\bibitem{chang2022maskgit}
Huiwen Chang, Han Zhang, Lu Jiang, Ce Liu, and William~T Freeman.
\newblock Maskgit: Masked generative image transformer.
\newblock In {\em Proceedings of the IEEE/CVF Conference on Computer Vision and
  Pattern Recognition}, pages 11315--11325, 2022.

\bibitem{chen2022pali}
Xi Chen, Xiao Wang, Soravit Changpinyo, AJ Piergiovanni, Piotr Padlewski,
  Daniel Salz, Sebastian Goodman, Adam Grycner, Basil Mustafa, Lucas Beyer,
  et~al.
\newblock Pali: A jointly-scaled multilingual language-image model.
\newblock {\em arXiv preprint arXiv:2209.06794}, 2022.

\bibitem{chen2019uniter}
Yen-Chun Chen, Linjie Li, Licheng Yu, Ahmed El~Kholy, Faisal Ahmed, Zhe Gan, Yu
  Cheng, and Jingjing Liu.
\newblock Uniter: Learning universal image-text representations.
\newblock 2019.

\bibitem{child2019generating}
Rewon Child, Scott Gray, Alec Radford, and Ilya Sutskever.
\newblock Generating long sequences with sparse transformers.
\newblock {\em arXiv preprint arXiv:1904.10509}, 2019.

\bibitem{dai2022enabling}
Wenliang Dai, Lu Hou, Lifeng Shang, Xin Jiang, Qun Liu, and Pascale Fung.
\newblock Enabling multimodal generation on clip via vision-language knowledge
  distillation.
\newblock {\em arXiv preprint arXiv:2203.06386}, 2022.

\bibitem{devlin2018bert}
Jacob Devlin, Ming-Wei Chang, Kenton Lee, and Kristina Toutanova.
\newblock Bert: Pre-training of deep bidirectional transformers for language
  understanding.
\newblock {\em arXiv preprint arXiv:1810.04805}, 2018.

\bibitem{ding2021cogview}
Ming Ding, Zhuoyi Yang, Wenyi Hong, Wendi Zheng, Chang Zhou, Da Yin, Junyang
  Lin, Xu Zou, Zhou Shao, Hongxia Yang, et~al.
\newblock Cogview: Mastering text-to-image generation via transformers.
\newblock {\em Advances in Neural Information Processing Systems},
  34:19822--19835, 2021.

\bibitem{ding2022cogview2}
Ming Ding, Wendi Zheng, Wenyi Hong, and Jie Tang.
\newblock Cogview2: Faster and better text-to-image generation via hierarchical
  transformers.
\newblock {\em arXiv preprint arXiv:2204.14217}, 2022.

\bibitem{dong2019unified}
Li Dong, Nan Yang, Wenhui Wang, Furu Wei, Xiaodong Liu, Yu Wang, Jianfeng Gao,
  Ming Zhou, and Hsiao-Wuen Hon.
\newblock Unified language model pre-training for natural language
  understanding and generation.
\newblock {\em Advances in neural information processing systems}, 32, 2019.

\bibitem{dosovitskiy2020image}
Alexey Dosovitskiy, Lucas Beyer, Alexander Kolesnikov, Dirk Weissenborn,
  Xiaohua Zhai, Thomas Unterthiner, Mostafa Dehghani, Matthias Minderer, Georg
  Heigold, Sylvain Gelly, et~al.
\newblock An image is worth 16x16 words: Transformers for image recognition at
  scale.
\newblock {\em arXiv preprint arXiv:2010.11929}, 2020.

\bibitem{gafni2022make}
Oran Gafni, Adam Polyak, Oron Ashual, Shelly Sheynin, Devi Parikh, and Yaniv
  Taigman.
\newblock Make-a-scene: Scene-based text-to-image generation with human priors.
\newblock In {\em Computer Vision--ECCV 2022: 17th European Conference, Tel
  Aviv, Israel, October 23--27, 2022, Proceedings, Part XV}, pages 89--106.
  Springer, 2022.

\bibitem{heusel2017gans}
Martin Heusel, Hubert Ramsauer, Thomas Unterthiner, Bernhard Nessler, and Sepp
  Hochreiter.
\newblock Gans trained by a two time-scale update rule converge to a local nash
  equilibrium.
\newblock {\em Advances in neural information processing systems}, 30, 2017.

\bibitem{ho2022classifier}
Jonathan Ho and Tim Salimans.
\newblock Classifier-free diffusion guidance.
\newblock {\em arXiv preprint arXiv:2207.12598}, 2022.

\bibitem{huang2021unifying}
Yupan Huang, Hongwei Xue, Bei Liu, and Yutong Lu.
\newblock Unifying multimodal transformer for bi-directional image and text
  generation.
\newblock In {\em Proceedings of the 29th ACM International Conference on
  Multimedia}, pages 1138--1147, 2021.

\bibitem{hutchinson-etal-2022-underspecification}
Ben Hutchinson, Jason Baldridge, and Vinodkumar Prabhakaran.
\newblock Underspecification in scene description-to-depiction tasks.
\newblock In {\em Proceedings of the 2nd Conference of the Asia-Pacific Chapter
  of the Association for Computational Linguistics and the 12th International
  Joint Conference on Natural Language Processing (Volume 1: Long Papers)},
  pages 1172--1184, Online only, Nov. 2022. Association for Computational
  Linguistics.

\bibitem{jia2021scaling}
Chao Jia, Yinfei Yang, Ye Xia, Yi-Ting Chen, Zarana Parekh, Hieu Pham, Quoc Le,
  Yun-Hsuan Sung, Zhen Li, and Tom Duerig.
\newblock Scaling up visual and vision-language representation learning with
  noisy text supervision.
\newblock In {\em International Conference on Machine Learning}, pages
  4904--4916. PMLR, 2021.

\bibitem{kim2022verse}
Taehoon Kim, Gwangmo Song, Sihaeng Lee, Sangyun Kim, Yewon Seo, Soonyoung Lee,
  Seung~Hwan Kim, Honglak Lee, and Kyunghoon Bae.
\newblock L-verse: Bidirectional generation between image and text.
\newblock In {\em Proceedings of the IEEE/CVF Conference on Computer Vision and
  Pattern Recognition}, pages 16526--16536, 2022.

\bibitem{li2022blip}
Junnan Li, Dongxu Li, Caiming Xiong, and Steven Hoi.
\newblock Blip: Bootstrapping language-image pre-training for unified
  vision-language understanding and generation.
\newblock In {\em International Conference on Machine Learning}, pages
  12888--12900. PMLR, 2022.

\bibitem{li2021align}
Junnan Li, Ramprasaath Selvaraju, Akhilesh Gotmare, Shafiq Joty, Caiming Xiong,
  and Steven Chu~Hong Hoi.
\newblock Align before fuse: Vision and language representation learning with
  momentum distillation.
\newblock {\em Advances in neural information processing systems},
  34:9694--9705, 2021.

\bibitem{li2019visualbert}
Liunian~Harold Li, Mark Yatskar, Da Yin, Cho-Jui Hsieh, and Kai-Wei Chang.
\newblock Visualbert: A simple and performant baseline for vision and language.
\newblock {\em arXiv preprint arXiv:1908.03557}, 2019.

\bibitem{lin2014microsoft}
Tsung-Yi Lin, Michael Maire, Serge Belongie, James Hays, Pietro Perona, Deva
  Ramanan, Piotr Doll{\'a}r, and C~Lawrence Zitnick.
\newblock Microsoft coco: Common objects in context.
\newblock In {\em Computer Vision--ECCV 2014: 13th European Conference, Zurich,
  Switzerland, September 6-12, 2014, Proceedings, Part V 13}, pages 740--755.
  Springer, 2014.

\bibitem{liu2021swin}
Ze Liu, Yutong Lin, Yue Cao, Han Hu, Yixuan Wei, Zheng Zhang, Stephen Lin, and
  Baining Guo.
\newblock Swin transformer: Hierarchical vision transformer using shifted
  windows.
\newblock In {\em Proceedings of the IEEE/CVF international conference on
  computer vision}, pages 10012--10022, 2021.

\bibitem{lu2022unified}
Jiasen Lu, Christopher Clark, Rowan Zellers, Roozbeh Mottaghi, and Aniruddha
  Kembhavi.
\newblock Unified-io: A unified model for vision, language, and multi-modal
  tasks.
\newblock {\em arXiv preprint arXiv:2206.08916}, 2022.

\bibitem{mu2022slip}
Norman Mu, Alexander Kirillov, David Wagner, and Saining Xie.
\newblock Slip: Self-supervision meets language-image pre-training.
\newblock In {\em Computer Vision--ECCV 2022: 17th European Conference, Tel
  Aviv, Israel, October 23--27, 2022, Proceedings, Part XXVI}, pages 529--544.
  Springer, 2022.

\bibitem{pham2021combined}
Hieu Pham, Zihang Dai, Golnaz Ghiasi, Kenji Kawaguchi, Hanxiao Liu, Adams~Wei
  Yu, Jiahui Yu, Yi-Ting Chen, Minh-Thang Luong, Yonghui Wu, et~al.
\newblock Combined scaling for open-vocabulary image classification.
\newblock {\em arXiv e-prints}, pages arXiv--2111, 2021.

\bibitem{plummer2015flickr30k}
Bryan~A Plummer, Liwei Wang, Chris~M Cervantes, Juan~C Caicedo, Julia
  Hockenmaier, and Svetlana Lazebnik.
\newblock Flickr30k entities: Collecting region-to-phrase correspondences for
  richer image-to-sentence models.
\newblock In {\em Proceedings of the IEEE international conference on computer
  vision}, pages 2641--2649, 2015.

\bibitem{radford2021learning}
Alec Radford, Jong~Wook Kim, Chris Hallacy, Aditya Ramesh, Gabriel Goh,
  Sandhini Agarwal, Girish Sastry, Amanda Askell, Pamela Mishkin, Jack Clark,
  et~al.
\newblock Learning transferable visual models from natural language
  supervision.
\newblock In {\em International conference on machine learning}, pages
  8748--8763. PMLR, 2021.

\bibitem{radford2019language}
Alec Radford, Jeffrey Wu, Rewon Child, David Luan, Dario Amodei, Ilya
  Sutskever, et~al.
\newblock Language models are unsupervised multitask learners.
\newblock {\em OpenAI blog}, 1(8):9, 2019.

\bibitem{raffel2020exploring}
Colin Raffel, Noam Shazeer, Adam Roberts, Katherine Lee, Sharan Narang, Michael
  Matena, Yanqi Zhou, Wei Li, and Peter~J Liu.
\newblock Exploring the limits of transfer learning with a unified text-to-text
  transformer.
\newblock {\em The Journal of Machine Learning Research}, 21(1):5485--5551,
  2020.

\bibitem{ramesh2022hierarchical}
Aditya Ramesh, Prafulla Dhariwal, Alex Nichol, Casey Chu, and Mark Chen.
\newblock Hierarchical text-conditional image generation with clip latents.
\newblock {\em arXiv preprint arXiv:2204.06125}, 2022.

\bibitem{ramesh2021zero}
Aditya Ramesh, Mikhail Pavlov, Gabriel Goh, Scott Gray, Chelsea Voss, Alec
  Radford, Mark Chen, and Ilya Sutskever.
\newblock Zero-shot text-to-image generation.
\newblock In {\em International Conference on Machine Learning}, pages
  8821--8831. PMLR, 2021.

\bibitem{rombach2022high}
Robin Rombach, Andreas Blattmann, Dominik Lorenz, Patrick Esser, and Bj{\"o}rn
  Ommer.
\newblock High-resolution image synthesis with latent diffusion models.
\newblock In {\em Proceedings of the IEEE/CVF Conference on Computer Vision and
  Pattern Recognition}, pages 10684--10695, 2022.

\bibitem{russakovsky2015imagenet}
Olga Russakovsky, Jia Deng, Hao Su, Jonathan Krause, Sanjeev Satheesh, Sean Ma,
  Zhiheng Huang, Andrej Karpathy, Aditya Khosla, Michael Bernstein, et~al.
\newblock Imagenet large scale visual recognition challenge.
\newblock {\em International journal of computer vision}, 115:211--252, 2015.

\bibitem{sahak2023denoising}
Hshmat Sahak, Daniel Watson, Chitwan Saharia, and David Fleet.
\newblock Denoising diffusion probabilistic models for robust image
  super-resolution in the wild.
\newblock {\em arXiv preprint arXiv:2302.07864}, 2023.

\bibitem{saharia2022photorealistic}
Chitwan Saharia, William Chan, Saurabh Saxena, Lala Li, Jay Whang, Emily
  Denton, Seyed Kamyar~Seyed Ghasemipour, Burcu~Karagol Ayan, S~Sara Mahdavi,
  Rapha~Gontijo Lopes, et~al.
\newblock Photorealistic text-to-image diffusion models with deep language
  understanding.
\newblock {\em arXiv preprint arXiv:2205.11487}, 2022.

\bibitem{shazeer2018adafactor}
Noam Shazeer and Mitchell Stern.
\newblock Adafactor: Adaptive learning rates with sublinear memory cost.
\newblock In {\em International Conference on Machine Learning}, pages
  4596--4604. PMLR, 2018.

\bibitem{shen2021much}
Sheng Shen, Liunian~Harold Li, Hao Tan, Mohit Bansal, Anna Rohrbach, Kai-Wei
  Chang, Zhewei Yao, and Kurt Keutzer.
\newblock How much can clip benefit vision-and-language tasks?
\newblock {\em arXiv preprint arXiv:2107.06383}, 2021.

\bibitem{Pax}
Pax Team.
\newblock Paxml, aka pax, https://github.com/google/paxml, 2023.

\bibitem{tewel2021zero}
Yoad Tewel, Yoav Shalev, Idan Schwartz, and Lior Wolf.
\newblock Zero-shot image-to-text generation for visual-semantic arithmetic.
\newblock {\em arXiv preprint arXiv:2111.14447}, 2021.

\bibitem{touvron2021training}
Hugo Touvron, Matthieu Cord, Matthijs Douze, Francisco Massa, Alexandre
  Sablayrolles, and Herv{\'e} J{\'e}gou.
\newblock Training data-efficient image transformers \& distillation through
  attention.
\newblock In {\em International conference on machine learning}, pages
  10347--10357. PMLR, 2021.

\bibitem{tschannen2022image}
Michael Tschannen, Basil Mustafa, and Neil Houlsby.
\newblock Image-and-language understanding from pixels only.
\newblock {\em arXiv preprint arXiv:2212.08045}, 2022.

\bibitem{NIPS2017_7a98af17}
Aaron van~den Oord, Oriol Vinyals, and koray kavukcuoglu.
\newblock Neural discrete representation learning.
\newblock In I. Guyon, U.~Von Luxburg, S. Bengio, H. Wallach, R. Fergus, S.
  Vishwanathan, and R. Garnett, editors, {\em Advances in Neural Information
  Processing Systems}, volume~30. Curran Associates, Inc., 2017.

\bibitem{vaswani2017attention}
Ashish Vaswani, Noam Shazeer, Niki Parmar, Jakob Uszkoreit, Llion Jones,
  Aidan~N Gomez, {\L}ukasz Kaiser, and Illia Polosukhin.
\newblock Attention is all you need.
\newblock {\em Advances in neural information processing systems}, 30, 2017.

\bibitem{wang2022git}
Jianfeng Wang, Zhengyuan Yang, Xiaowei Hu, Linjie Li, Kevin Lin, Zhe Gan,
  Zicheng Liu, Ce Liu, and Lijuan Wang.
\newblock Git: A generative image-to-text transformer for vision and language.
\newblock {\em arXiv preprint arXiv:2205.14100}, 2022.

\bibitem{wang2022ofa}
Peng Wang, An Yang, Rui Men, Junyang Lin, Shuai Bai, Zhikang Li, Jianxin Ma,
  Chang Zhou, Jingren Zhou, and Hongxia Yang.
\newblock Ofa: Unifying architectures, tasks, and modalities through a simple
  sequence-to-sequence learning framework.
\newblock In {\em International Conference on Machine Learning}, pages
  23318--23340. PMLR, 2022.

\bibitem{wang2022image}
Wenhui Wang, Hangbo Bao, Li Dong, Johan Bjorck, Zhiliang Peng, Qiang Liu, Kriti
  Aggarwal, Owais~Khan Mohammed, Saksham Singhal, Subhojit Som, et~al.
\newblock Image as a foreign language: Beit pretraining for all vision and
  vision-language tasks.
\newblock {\em arXiv preprint arXiv:2208.10442}, 2022.

\bibitem{wang2021simvlm}
Zirui Wang, Jiahui Yu, Adams~Wei Yu, Zihang Dai, Yulia Tsvetkov, and Yuan Cao.
\newblock Simvlm: Simple visual language model pretraining with weak
  supervision.
\newblock {\em arXiv preprint arXiv:2108.10904}, 2021.

\bibitem{xie2019visual}
Ning Xie, Farley Lai, Derek Doran, and Asim Kadav.
\newblock Visual entailment: A novel task for fine-grained image understanding.
\newblock {\em arXiv preprint arXiv:1901.06706}, 2019.

\bibitem{yao2021filip}
Lewei Yao, Runhui Huang, Lu Hou, Guansong Lu, Minzhe Niu, Hang Xu, Xiaodan
  Liang, Zhenguo Li, Xin Jiang, and Chunjing Xu.
\newblock Filip: fine-grained interactive language-image pre-training.
\newblock {\em arXiv preprint arXiv:2111.07783}, 2021.

\bibitem{you2022learning}
Haoxuan You, Luowei Zhou, Bin Xiao, Noel Codella, Yu Cheng, Ruochen Xu, Shih-Fu
  Chang, and Lu Yuan.
\newblock Learning visual representation from modality-shared contrastive
  language-image pre-training.
\newblock In {\em Computer Vision--ECCV 2022: 17th European Conference, Tel
  Aviv, Israel, October 23--27, 2022, Proceedings, Part XXVII}, pages 69--87.
  Springer, 2022.

\bibitem{yu2021vector}
Jiahui Yu, Xin Li, Jing~Yu Koh, Han Zhang, Ruoming Pang, James Qin, Alexander
  Ku, Yuanzhong Xu, Jason Baldridge, and Yonghui Wu.
\newblock Vector-quantized image modeling with improved vqgan.
\newblock {\em arXiv preprint arXiv:2110.04627}, 2021.

\bibitem{yu2022coca}
Jiahui Yu, Zirui Wang, Vijay Vasudevan, Legg Yeung, Mojtaba Seyedhosseini, and
  Yonghui Wu.
\newblock Coca: Contrastive captioners are image-text foundation models.
\newblock {\em arXiv preprint arXiv:2205.01917}, 2022.

\bibitem{yu2022scaling}
Jiahui Yu, Yuanzhong Xu, Jing~Yu Koh, Thang Luong, Gunjan Baid, Zirui Wang,
  Vijay Vasudevan, Alexander Ku, Yinfei Yang, Burcu~Karagol Ayan, et~al.
\newblock Scaling autoregressive models for content-rich text-to-image
  generation.
\newblock {\em arXiv preprint arXiv:2206.10789}, 2022.

\bibitem{yuan2021florence}
Lu Yuan, Dongdong Chen, Yi-Ling Chen, Noel Codella, Xiyang Dai, Jianfeng Gao,
  Houdong Hu, Xuedong Huang, Boxin Li, Chunyuan Li, et~al.
\newblock Florence: A new foundation model for computer vision.
\newblock {\em arXiv preprint arXiv:2111.11432}, 2021.

\bibitem{zhai2022scaling}
Xiaohua Zhai, Alexander Kolesnikov, Neil Houlsby, and Lucas Beyer.
\newblock Scaling vision transformers.
\newblock In {\em Proceedings of the IEEE/CVF Conference on Computer Vision and
  Pattern Recognition}, pages 12104--12113, 2022.

\bibitem{zhai2022lit}
Xiaohua Zhai, Xiao Wang, Basil Mustafa, Andreas Steiner, Daniel Keysers,
  Alexander Kolesnikov, and Lucas Beyer.
\newblock Lit: Zero-shot transfer with locked-image text tuning.
\newblock In {\em Proceedings of the IEEE/CVF Conference on Computer Vision and
  Pattern Recognition}, pages 18123--18133, 2022.

\bibitem{zhang2021ernie}
Han Zhang, Weichong Yin, Yewei Fang, Lanxin Li, Boqiang Duan, Zhihua Wu, Yu
  Sun, Hao Tian, Hua Wu, and Haifeng Wang.
\newblock Ernie-vilg: Unified generative pre-training for bidirectional
  vision-language generation.
\newblock {\em arXiv preprint arXiv:2112.15283}, 2021.

\bibitem{zhang2021vinvl}
Pengchuan Zhang, Xiujun Li, Xiaowei Hu, Jianwei Yang, Lei Zhang, Lijuan Wang,
  Yejin Choi, and Jianfeng Gao.
\newblock Vinvl: Revisiting visual representations in vision-language models.
\newblock In {\em Proceedings of the IEEE/CVF Conference on Computer Vision and
  Pattern Recognition}, pages 5579--5588, 2021.

\bibitem{zhou2020unified}
Luowei Zhou, Hamid Palangi, Lei Zhang, Houdong Hu, Jason Corso, and Jianfeng
  Gao.
\newblock Unified vision-language pre-training for image captioning and vqa.
\newblock In {\em Proceedings of the AAAI conference on artificial
  intelligence}, volume~34, pages 13041--13049, 2020.

\end{thebibliography}
}


\begin{table*}[h]
\small
\setlength\tabcolsep{10pt} %
\renewcommand{\arraystretch}{1.2} %
\centering
        \begin{tabular}{l|c|c|c|c|c}
        \toprule
         & \multirow{2}{*}{\makecell[c]{ImageNet \\ Linear Probe}} & \multirow{2}{*}{VQA} & \multirow{2}{*}{SNLI-VE} & \multirow{2}{*}{\makecell[c]{MS-COCO \\ Captioning}}   & \multirow{2}{*}{\makecell[c]{MS-COCO  \\  T2I Generation}} \\
          &  &  &  &  & \\
         \hline
         Optimizer &  SGD & \multicolumn{4}{c}{Adafacter}\\
         \hline
         Gradient Clip &  \multicolumn{5}{c}{1.0}\\
         \hline
         LR decay  schedule &  Cosine Schedule to zero & \multicolumn{3}{c|}{Linear Schedule to zero}  & Exponential Schedule to zero\\
         \hline
         RandAugment & 2, 5 & \multicolumn{3}{c|}{1, 10}  &  None\\
         \hline
         Training Step &  225k & 100k & 50k & 15k & 100k\\
         \hline
         Warm-Up Step &  0 & 1000 &  1000 & 500 & 1000\\
         \hline
         Batch Size &  512 & 32 & 128 & 128 & 256\\
         \hline
         Learning Rate &  3.2 & 1e-5 & 5e-5 & 5e-6 & 1e-5\\
         \hline
         Weight Decay &  0.0 & 0.1 & 0.1 & 0.01 & 0.045\\
        \bottomrule
        \end{tabular}
\caption{Hyper-parameters used in the multimodal experiments.}
\label{tab:hps}
\end{table*}

\section{Appendix}

\subsection{Ablation on Loss Weight}

\begin{figure*}[t]
\centering
  \includegraphics[width=1\linewidth]{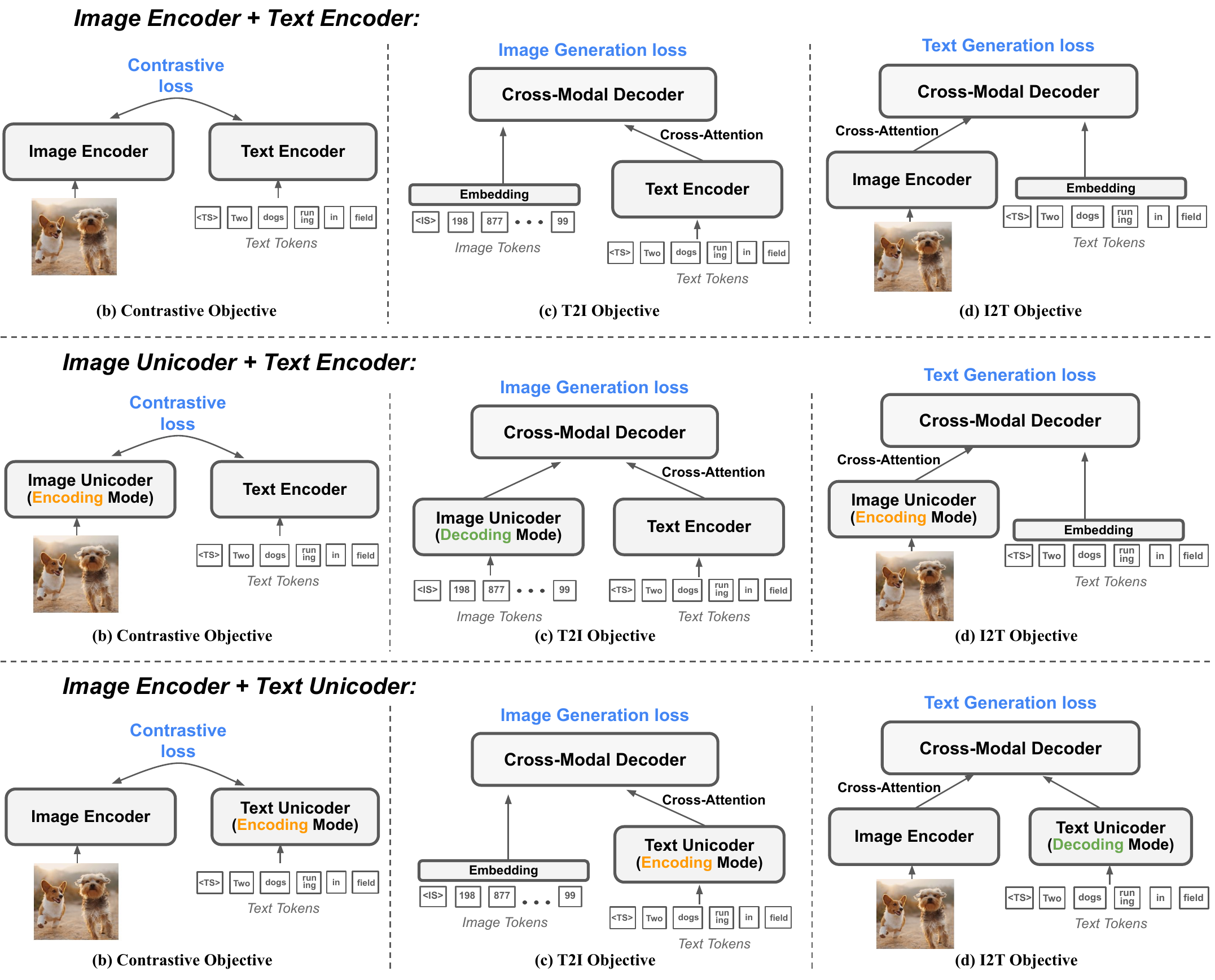}
  \caption{Diagram of three compared models in the ablation of \textbf{unicoder vs. encoder}. \textbf{Top}: Replacing both image unicoder and text unicoder with image encoder and text encoder respectively.  \textbf{Middle}: Replacing text unicoder with text encoder while keeping image unicoder. \textbf{Bottom}: Replacing image unicoder with image encoder while keeping text unicoder. }
  \label{fig:diagram_supp}
\end{figure*}

Given three objectives, we ablate different weights for them and select the best one as the default configuration for all experiments.  We start with T2I and I2T first: given the weight of T2I fixed to 1, we ablate the loss of I2T. Then given T2I and I2T loss both fixed, the weight of contrastive loss is ablated. As we can see in Tab. \ref{tab:exp_ablate_weight}, a high weight of I2T such as 1 will hurt the image generation heavily but also improve VQA. On the other hand, a high weight of contrastive loss like 0.4 will not essentially improve image recognition and hurts both VQA and image generation. Overall, we chose Con.:T2I:I2T = 0.1:0.2:1 as our default setting, as it achieves a good trade-off between three losses. 

\begin{table}[h]
\caption{Ablation on weights of three losses. Con. denotes contrastive loss. T2I denotes text-to-image generation loss. I2T denotes image-to-text generation loss. ZS IN. denotes zero-shot ImageNet classification. ZS IG. denotes zero-shot text-to-image generation on MS-COCO, which is evaluated by FID and lower FID is better. }
\label{tab:exp_ablate_weight}
\begin{center}
\begin{tabular}{cccccc}
\toprule
 \multicolumn{3}{c}{Weights} & \multicolumn{3}{c}{Evaluation} \\
 \cmidrule(lr){1-3}  \cmidrule(lr){4-6}
 Con. & T2I & I2T & ZS IN. & VQA & ZS IG. ($\downarrow$) \\
 \midrule
 - &  0.1 & 1  & - & 63.2 & 13.17 \\
 - & 0.2 & 1  & - & 65.4 & 13.24\\
 - & 1 & 1 & - & 67.8 & 16.33 \\
\midrule{}
 0.1 &  0.2 & 1  & 71.1 & 66.9 & 13.31 \\
 0.4 & 0.2 & 1 & 71.2 & 66.5 & 13.92 \\
\midrule
\bottomrule
\end{tabular}
\end{center}
\end{table}

\subsection{Hyperparameters in Fine-tuning}
In Tab. \ref{tab:hps}, we present the hyperparameters we used in fine-tuning/linear probing of \cocadraw{}.

\begin{figure*}[t]
 \centering
  \includegraphics[width=0.88\linewidth]{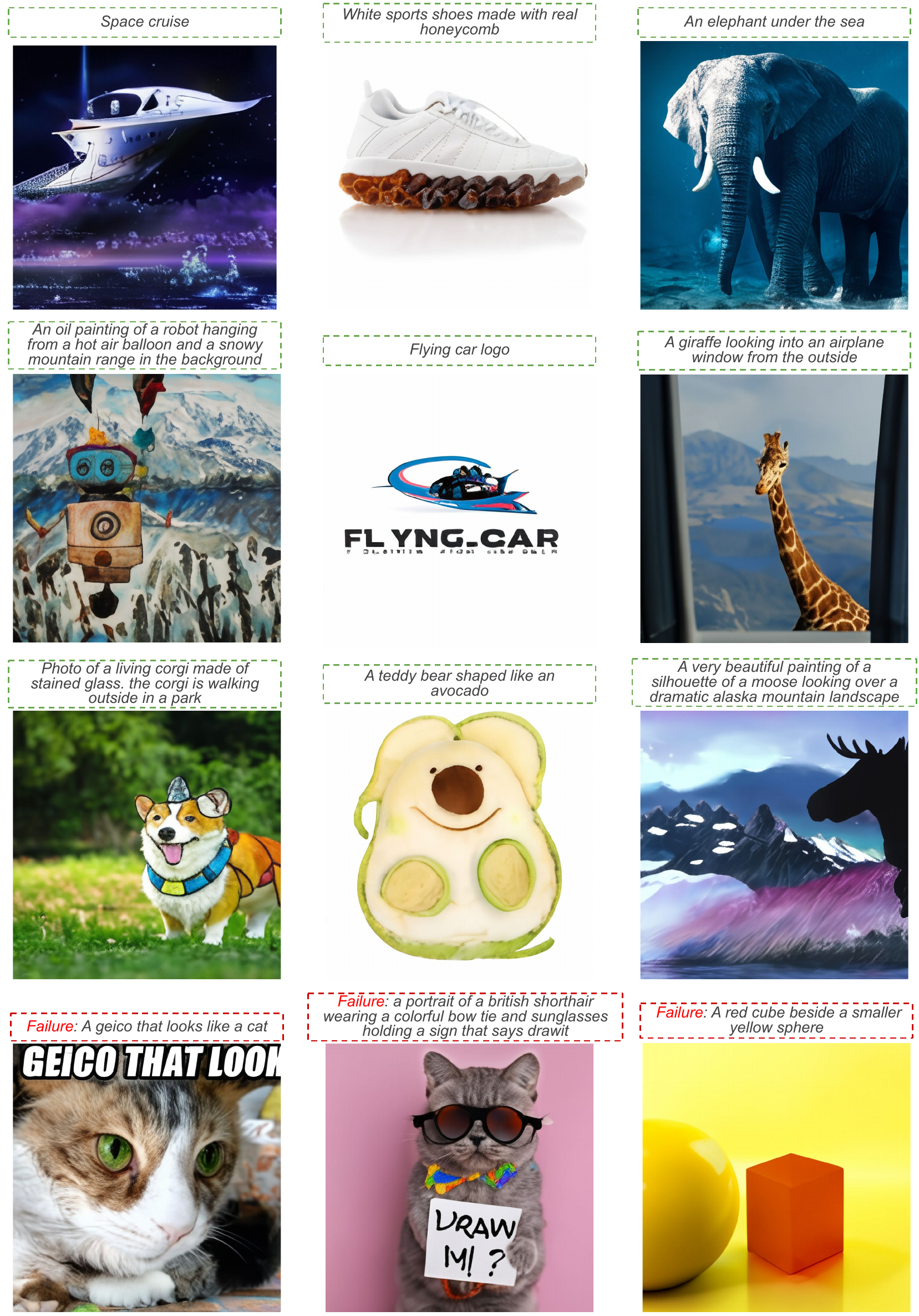}
  \caption{More qualitative results of zero-shot text-to-image generation from \cocadrawlarge{} with both good and failed cases. }
  \label{fig:vis_supp}
\end{figure*}

\subsection{Illustration of Replacing Unicoders with Encoders in \cocadraw} 
In Sec.4.4, we ablate \textbf{Unicoder  vs.   Encoder} and demonstrate the effectiveness of proposed unicoders. In Fig. \ref{fig:diagram_supp} we show the diagram of using image and text encoders, image encoder+text unicoder, and image unicoder+text encoder.  As we can see, encoders can only encode visual or textual features while unicoders can perform both encoding and decoding, which shares the knowledge and boosts the generation result as shown in previous ablation. It's noted that, replacing uncoders with encoders doesn't save the parameters, because inside unicoders, encoding and decoding share the same Transformer parameters. So in terms of parameter efficiency, unicoder and encoder have no difference. 

\subsection{More visualization} 
In Fig. \ref{fig:vis_supp}, We attach more visualization of \cocadrawlarge{} on zero-shot text-to-image generation with novel prompts in PartiPrompts \cite{yu2022scaling}. For better visualization when zoom-in, we employ \cite{sahak2023denoising} as the super-resolution module to upsample generated 256x256 images to 1024x1024 images. It's noted that when computing FID, we still use 256x256 images and the high-resolution ones are only used for visualization. In failed cases, we find that: (1) \cocadraw{} sometimes messes up the size attributes of two objects. For example, in the last example, yellow sphere ought to be smaller. (2) \cocadraw{} sometimes couldn't render the details of words in text very well. In the second last example, ``DRAWIT'' is rendered as ``DRAWMI?''. (3) \cocadraw{} occasionally misunderstands the text. In the third last example, we expect a geico that looks like a cat whereas \cocadraw{} first renders ``GEICO THAT LOOK'' then generates a cat. It's indeed a new way to interpret the text but not the desired way of humans.

\end{document}